\documentclass{article}
\usepackage[T1]{fontenc} 
\usepackage[utf8]{inputenc} 
\usepackage{ismir,amsmath,cite,url}
\usepackage{graphicx}
\usepackage{color}
\usepackage{algorithm}
\usepackage{amssymb}
\usepackage[noend]{algpseudocode}
\usepackage{maths}
\usepackage{multirow}

\title{Ultra-light deep MIR by trimming lottery tickets}

\oneauthor
{Philippe Esling, Theis Bazin, Adrien Bitton, Tristan Carsault, Ninon Devis}
{{IRCAM - Sorbonne Université, CNRS UMR 9912 - 1, place Igor Stravinsky, Paris, France} \\
{\tt {esling@ircam.fr}}
}

\begin{document}

\maketitle
\begin{abstract}

Current state-of-the-art results in Music Information Retrieval are largely dominated by deep learning approaches. These provide unprecedented accuracy across all tasks. However, the consistently overlooked downside of these models is their stunningly massive complexity, which seems concomitantly crucial to their success. 

In this paper, we address this issue by proposing a model pruning method based on the \textit{lottery ticket} hypothesis. We modify the original approach to allow for explicitly removing parameters, through \textit{structured trimming} of entire units, instead of simply masking individual weights. This leads to models which are effectively lighter in terms of size, memory and number of operations. 

We show that our proposal can remove up to 90\% of the model parameters without loss of accuracy, leading to ultra-light deep MIR models. We confirm the surprising result that, at smaller compression ratios (removing up to 85\% of a network), lighter models consistently outperform their heavier counterparts. We exhibit these results on a large array of MIR tasks including \textit{audio classification}, \textit{pitch recognition}, \textit{chord extraction}, \textit{drum transcription} and \textit{onset estimation}. The resulting ultra-light deep learning models for MIR can run on CPU, and can even fit on embedded devices with minimal degradation of accuracy. \footnote{Supplementary results and code to reproduce experiments are available at \href{https://github.com/acids-ircam/lottery\_mir}{https://github.com/acids-ircam/lottery\_mir}}

\end{abstract}
\section{Introduction}\label{sec:introduction}

Over the past decades, Music Information Retrieval (MIR) has witnessed a growing interest, with a wide variety of tasks such as \textit{genre classification}, \textit{chord extraction} and \textit{music recommendation} \cite{casey2008content} being increasingly implemented in end-user products. Recently, MIR has predominantly improved with machine learning, and almost all state-of-art accuracies are obtained by deep learning models \cite{humphrey2012moving}. Although these approaches provide unprecedented results, the major issue in modern deep learning lies in the tremendous complexity and immense size of the models employed. Indeed, deep networks for images can reach up to billions of parameters and new leaps in accuracy seem to only come by worsening this situation. As a showering example of this complexity \cite{you2017imagenet}, the inference on a single image in the pervasive ResNet model \cite{he2016deep} requires 7.7 GFLOPS\footnote{FLOPS: floating point operations}. This exploding size leads to profound issues in both the use and understanding of these models. As they are extremely demanding in computation and memory, it precludes their implementation in end-user embedded systems which prevails in audio applications, and also raises some serious environmental issues. Finally, such complexity decreases the potential interpretability of these models. 

\begin{figure}
 \includegraphics[width=.48\textwidth]{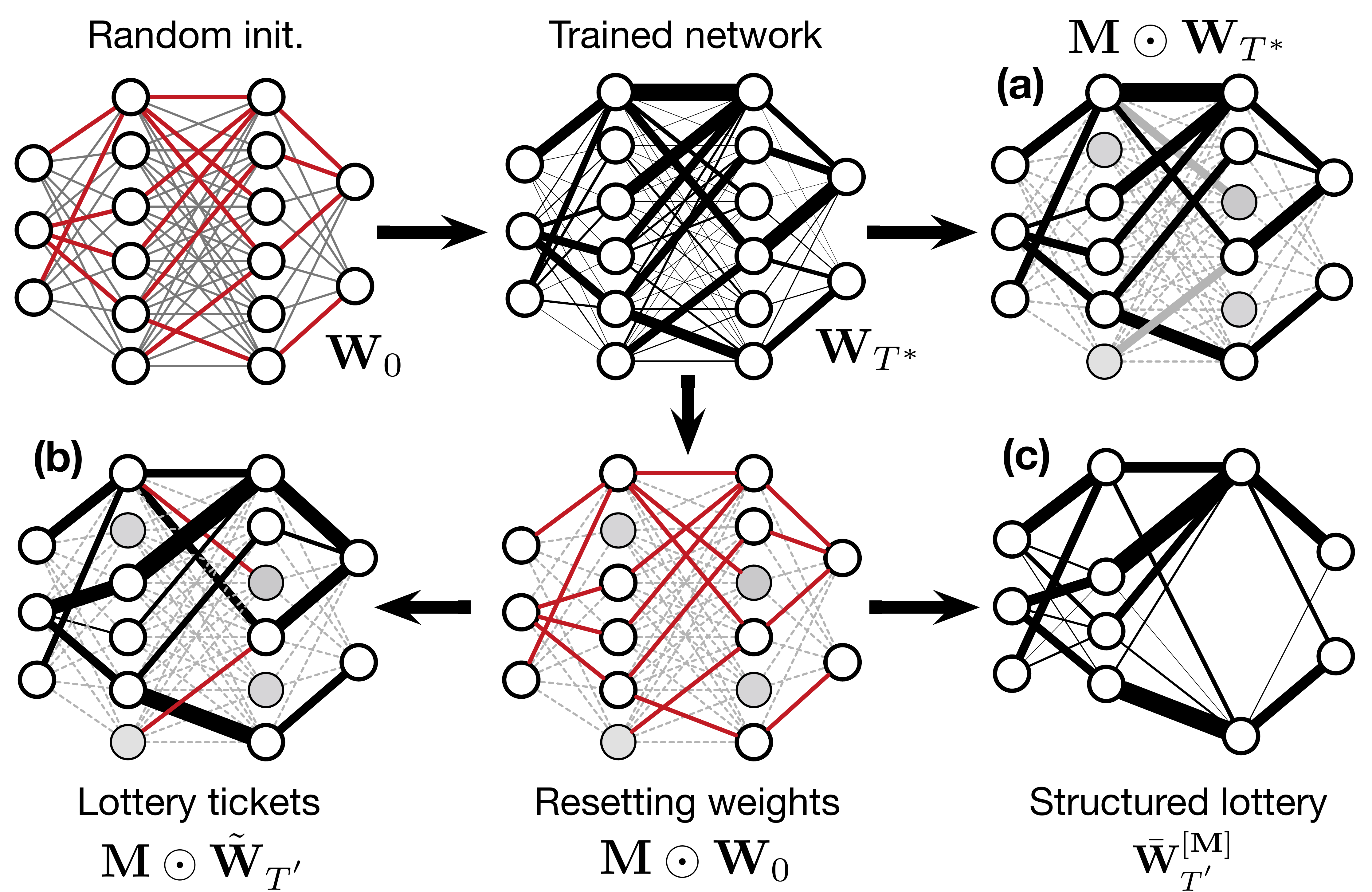}
 \caption{Comparing (a) traditional pruning with (b) the lottery ticket hypothesis, and (c) our structured lottery approach to obtain ultra-light deep networks.}
 \label{fig:lottery_compare}
\end{figure}

The idea of eliminating unnecessary weights (\textit{pruning}) was proposed early for neural networks \cite{lecun1990optimal}. Most methods are based on masking the smallest-amplitude weights from a large network, as depicted in \figref{fig:lottery_compare}. Other approaches such as \textit{quantization} \cite{han2015deep} or \textit{knowledge distillation} \cite{hinton2015distilling} have been proposed to decrease the size and energy consumption of trained models with equivalent accuracy. However, keeping the original accuracy of complex large models seems only possible at low compression rates \cite{liu2018rethinking}. Furthermore, recent benchmark studies \cite{gale2019state} pointed out that most of the proposed methods seems to achieve a similar efficiency in both accuracy and model size.

Recently, the \textit{lottery ticket hypothesis} \cite{frankle2018lottery} suggested that randomly-initialized neural networks already contain powerful subnetworks (called \textit{winning tickets}) that could reach the same or higher accuracy than the original networks if they were trained in isolation. Hence, by finding these subnetworks, we could drastically prune most of the weights in large networks and still obtain the same level of accuracy. This implies that the same task could be solved in a very lightweight, memory and energy-efficient way. Furthermore, these subnetworks could be easier to analyze, which could simplify further works towards explainability \cite{frankle2019linear}. Several studies have analyzed different properties of this hypothesis \cite{frankle2019lottery, frankle2020the, morcos2019one}, as it raises the exciting prospect to obtain much smaller networks that provide a similar accuracy compared to the typically larger state-of-art models. However, this method has two major flaws. First, it has a large training cost, as finding winning tickets seems to only be stable when the training is repeated multiple times over iteratively smaller networks \cite{you2019drawing}. Second, pruning is done by \textit{masking} the weights, which means that the resulting networks retain the size and computation cost of the original ones, even if most of the weights are unused.

In this paper, we extend the lottery approach to effectively remove weights, obtaining models with a lower size and inference time, while still maintaining a commensurate accuracy. To do so, we introduce a method based on the lottery ticket hypothesis, and we replace the masking operation with a structured pruning operation (termed \textit{trimming} here). The original network capacity is reduced by removing entire computation units (or convolutional channels). This alleviates issues of the original lottery ticket method as, although we still need to repeat the training, it becomes faster at each iteration. We discuss different criteria for selecting the units and their differences to the original lottery ticket hypothesis. Notably, unstructured masking allows to work on local connectivity patterns, whereas trimming can only impact this aspect if we perform \textit{global} selection (ranking units across the network). We show that this approach can be successfully applied across MIR tasks, leading to \textit{ultra-light} deep MIR models. We evaluate the efficiency of replacing the masking operation by our trimming criterion and show that we still obtain commensurate accuracy when removing up to 90\% of the model parameters. We also maintain the surprising result \cite{frankle2018lottery} that lighter models (removing up to 85\% of the network) obtain higher accuracy, while we effectively reduce the model size. 
We evaluate these results on a large array of MIR tasks including \textit{instrument} \cite{esling2018bridging} and \textit{singing voice classification} \cite{wilkins2018vocalset}, \textit{pitch recognition} \cite{kim2018crepe}, \textit{automatic chord extraction} \cite{carsault2018using}, \textit{drum transcription} \cite{choi2019deep} and \textit{onset estimation} \cite{schluter2014improved}. 

\section{State-of-art}
\label{sec:stateofart}

\subsection{Model compression and pruning}

Various approaches have been proposed for reducing the size of neural network models, while trying to maintain accuracy \cite{lecun1990optimal}. These approaches can be globally divided between \textit{pruning} or \textit{compressing} networks. We group in the \textit{compression} category the \textit{distillation} \cite{hinton2015distilling} (training a smaller model to fit the internal representations of a larger one) and \textit{quantization} \cite{han2015deep} (reducing the size of networks by using lower-resolution weights or binary numbers) approaches. Here, we focus on \textit{pruning}, but note that compression and quantization can be further applied on pruned models.

The goal of \textit{pruning} \cite{lecun1990optimal} is to identify and remove weights of a network that are not critical to its accuracy. The original approach to pruning starts by fitting a large and overparametrized network to completion. Then, we aim to mask the less relevant weights in this trained model based on a given selection method. This criterion tries to analyze the usefulness of different parameters, commonly based on their magnitude \cite{han2015learning}. Finally, the resulting masked network is \textit{fine-tuned}, trying to restore the accuracy of the original network \cite{gale2019state}. Hence, the critical aspect in this approach lies in the method of weight selection. This criterion can perform either a \textit{structured} or \textit{unstructured} and \textit{local} or \textit{global} selection. \textit{Unstructured} pruning acts on individual parameters separately, \textit{structured} pruning aims to effectively remove parts of the networks. Hence, unstructured methods are mostly based on \textit{masking} the weights based on their magnitude \cite{lecun1990optimal, han2015learning}. Oppositely, structured pruning aims to remove entire \textit{hidden units} or \textit{convolutional channels} from a network \cite{li2016pruning, liu2018rethinking}. 

However, recent studies showed that most pruning methods are mostly equivalent \cite{liu2018rethinking}. These approaches usually lead to smaller accuracy than the large network and at low pruning rates, with performance degrading with the amount of weights removed \cite{gale2019state}, although some are able to maintain (but not outperform) the original accuracy \cite{liu2018rethinking}.

\subsection{Lottery ticket hypothesis}

The recently proposed \textit{lottery ticket hypothesis} \cite{frankle2018lottery} states that inside a randomly-initialized network, there already exist some considerably smaller subnetworks which would be extremely efficient if trained in isolation. Hence, parts of the weights drawn by random initialization before training already provide a specific topology and parameter configuration that make training particularly effective. The major difference between this approach and the previous magnitude-based selection from which it is inspired \cite{han2015learning} is that the selected weights are \textit{reset} to their initialization value before retraining the smaller architecture. Doing so, very small subnetworks (less than 1\% of the original network size) could be found across several architectures, even outperforming the larger networks at smaller pruning ratios. For deeper architectures, they further showed \cite{frankle2019lottery} that winning tickets should be \textit{rewound} to a given iteration, rather than to initialization values. Interestingly, this seems to confirm that overparameterization is needed to find an optimal solution, but that a lighter solution exists, which is optimized in the compression phase of the training \cite{shwartz2017opening, frankle2020the}.


\subsubsection{Formalization}

We consider a network as a parametric function $f(\bx; \bW)$, with a set of weights $\bW\in\mathbb{R}^{D}$ that are first initialized through sampling $\bW_{0} \sim p(\bW)$. The weights are updated by using a training \textit{algorithm} $\Alg{i}{\bW_{0}}{}$ which maps initial weights $\bW_{0}$ to weights $\bW_i$ at iteration $i \in \{1,..,T\}$, by performing successive updates similar to
\begin{equation}
    \bW_{i+1}=\bW_{i} - \eta \nabla_{\bW} \mathcal{L}
\end{equation}
with a given loss function $\mathcal{L}$ and learning rate $\eta$.

A \textit{subnetwork} of the original network $f(\bx; \bW)$ can be defined as a tuple $(\bW, \bM)$ of the original weights $\bW \in \mathbb{R}^D$ and a mask $\mathbf{M} : \{0, 1\}^D$. Hence, the subnetwork computes the function $f(\bx; \bM \odot \bW)$, where $\odot$ denotes the element-wise product. 

\textbf{Lottery Ticket Hypothesis.} Given a randomly initialized network $f(\bx; \bW_{0})$ with $\bW_{0} \sim p(\bW)$, that is trained to reach accuracy $a^*$ in $T^*$ iterations, with final weights $\bW_{T^*}$, there exists a subnetwork $(\bW_k,\bM)$ with a given mask $\bM \in \{0, 1\}^{|\bW|}$ and iteration $k \ll T^{*}$, such that if we retrain this subnetwork, it will reach a \textit{commensurate accuracy} $a \geq a^*$ in \textit{commensurate iterations} $T \leq T^* - k$ and \textit{fewer parameters} $\lVert \bM \rVert_0 \ll |\bW|$.

These highly efficient subnetworks (called \textit{winning tickets}) depend on the original initialization, and can only be identified after full training \cite{frankle2019lottery}. Thus, the selected weights and remaining topology form the architecture of the winning ticket. These weights are \textit{reset} to their initialization values \emph{before} the network was trained or \textit{rewound} at an early iteration. The resulting architecture is then retrained until completion, and the whole process is repeated, as described in Algorithm~\ref{alg:lottery}.


\begin{algorithm}
    \caption{Lottery ticket training with rewinding}
    \label{alg:lottery}
    \begin{algorithmic}[1]
        \State $\bW_{0}\sim p(\bW)$          \Comment{Random initialization}
        \State $\bM = \bm{1}_{|\bW|}$        \Comment{Initial mask}
        \State $\bW_{k}=\Alg{k}{\bW_{0}\odot\bM}{}$         \Comment{Training for $k$ iterations}
        \While{$\mathcal{C}(\bM,a,\bW)$}     \Comment{Stopping criterion $\mathcal{C}$}
            \State $\bW_{T}=\Alg{T}{\bW_{k}\odot\bM}{}$         \Comment{Train until completion}
            \State $r = \mathcal{R}(\{\bW_{T^{*}}\})$      \Comment{Ranking criterion $\mathcal{R}$}
            \State $\bM = \mathcal{M}(r,\{\bW_{T^{*}}\})$    \Comment{Masking update $\mathcal{M}$}
        \EndWhile
    \end{algorithmic}
\end{algorithm}

In their original paper \cite{frankle2018lottery}, the authors underline the difference between \emph{one-shot} pruning (masking is applied all at once) and \emph{iterative pruning} (repeatedly pruning small parts of the network). They demonstrated that iterative pruning finds smaller architectures that reach higher accuracy than the original network and converge at earlier iterations. They showed on the MNIST dataset, that it was possible to keep the accuracy of large networks, even when masking up to $96.5\%$ of the weights. Their most intriguing result is that smaller networks consistently reach \textit{higher accuracy} than the original ones, even while removing up to $80\%$ of the weights. In a follow-up study \cite{frankle2019lottery}, they showed that these results could be obtained for deeper architectures, but only through the \textit{rewinding} operation.
Another exciting prospect of this hypothesis, is that the resulting subnetworks might encode implicit \textit{inductive biases} for a given task or type of data. In that case, winning tickets could be \textit{transferred} and trained on new tasks, even directly from their extremely lightweight versions \cite{morcos2019one}.


\subsection{Music Information Retrieval}

Music Information Retrieval (MIR) encompasses all tasks aimed at extracting high-level knowledge from music data. This field has witnessed a flourishing interest, with multiple tasks being increasingly tackled such as \textit{chord extraction}, \textit{drum transcription} and \textit{musical audio classification} \cite{casey2008content}. Originally, most MIR researches revolved around the idea of extracting a set of hand-crafted features from the signal (such as the Mel-Frequency Cepstral Coefficients), in order to use these as input to machine learning algorithms \cite{mcfee2012learning}. Feature-based techniques have been challenged by the advent of deep learning approaches \cite{humphrey2013feature}, which have shown impressive capacities to learn high-level features on complex data. They simultaneously set new state-of-art results across a wide range of MIR tasks, while opening the path towards unprecedented applications \cite{esling2020flow}. 

In this work, we consider a rather broad spectrum of MIR tasks where deep learning approaches are applied. Specifically, we address (i) \textit{audio classification} \cite{wilkins2018vocalset} (finding the class label of audio signals inside a predefined set), (ii) \textit{pitch recognition} \cite{kim2018crepe} (extracting the fundamental frequency of a monophonic audio recording), (iii) \textit{chord extraction} \cite{carsault2018using} (annotating audio with a given vocabulary of chords), (iv) \textit{onset estimation} \cite{schluter2014improved} (finding events in an audio stream) and (v) \textit{drum transcription} \cite{choi2019deep} (transforming drums audio signal into a score). We redirect interested readers to \cite{choi2017tutorial} for a comprehensive review.

One of the common denominator in deep learning methods applied across all MIR tasks is that their unprecedented accuracy comes at the expense of an increasing size and complexity. Indeed, deep networks for images now reach billions of parameters and leaps in accuracy seem to only come by worsening this situation. An example of this trend in MIR can be seen in the recently proposed CREPE model \cite{kim2018crepe} for pitch extraction. This task was largely handled through the YIN algorithm \cite{de2002yin}, an extremely simple algorithm, with few parameters and running with very low latency on CPU. For a modest gain in accuracy on the same task, CREPE requires \textit{22 million parameters}, 2.82 GFLOPS and 2.36 seconds on CPU to compute the pitch of a single 4-seconds sample. This exploding size leads to profound issues in both the use and understanding of these models. First, they are extremely demanding in energy consumption and memory, which precludes their implementation in end-user interfaces and also raises serious environmental issues. Furthermore, this complexity stands in the way of any potential interpretability of such models. 

\section{Methodology}

Here, we first discuss different selection criteria for structured network \textit{trimming}. Then, we discuss different normalization strategies that can allow to perform global selection of units across layers.  

\subsection{Trimming criteria}

In order to perform structured pruning, we need to evaluate the efficiency of \textit{entire units} of computation, rather than individual weights. In the case of convolutional networks, this would amount to analyze the \textit{channels} of each layer. Indeed, channel pruning appears more hardware friendly, and also allows to truly reduce the size of the final model. In the following definitions, we consider that any computation layer can be seen as a weighted transform $f(\bx, \bW^{(l)})$, with a matrix $\bW^{(l)} \in \R^{n_{out}\times n_{in}}$. Note that we intentionally simplify the notation for more complex layers (\textit{convolutional} or \textit{recurrent}), which embed more complicated matrices. However, we consider in the following that the selection criteria $\mathcal{C}(\bW^{(l)})$ is computed across the $n_{in}$ dimensions, and that it should produce a vector of $n_{out}$ dimensions. This vector is used to rank the usefulness of different computation units. Hence, after each training iteration, we replace the masking criterion by directly removing parts of the weight matrix for each layer
\begin{equation}
    \bW^{(l)} = \bW^{(l)}_{[C(\bW^{(l)}),C(\bW^{(l-1)})]}.
\end{equation}

Note that we need to carry the pruning criterion from the preceding layer $C(\bW^{(l-1)})$ in order to reflect potential changes in the structure of the network. All layers in the network that must maintain a given output dimensionality (such as the last layer) are defined as \textit{unprunable}. Following a similar approach than the lottery ticket hypothesis, the remaining weights in the resulting matrix $\bW^{l}$ are \textit{rewound} to their values from an earlier iteration \cite{frankle2020the}.

\textit{Magnitude.} 
We define a \textit{magnitude-based} criterion, similar to the original lottery \cite{frankle2018lottery}. However, we evaluate the overall magnitude of the weights for a complete unit as
\begin{equation}
    \mathcal{C}_{mag}(\bW^{(l)}_{i})= \sum_{j=1}^{N_{in}} \left| W^{(l)}_{i, j} \right|.
\end{equation}


\textit{Activation.}
We can rely on the activation statistics of each unit to analyze their importance. Hence, akin to the previous criterion, we perform a cumulative forward pass through the network after training the model and compute
\begin{equation}
    \mathcal{C}_{act}(\bW^{(l)}_{i})= \sum_{k=1}^{\mathcal{D}_{v}} \left| f(\bx_{k}, \bW^{(l)})_{i} \right|
\end{equation}
where we sum across examples of the validation set $\mathcal{D}_{v}$.

\textit{Normalization.}
An interesting direction proposed in \cite{you2019drawing} is to consider the \textit{scaling} factor $\gamma$ in batch normalization layers to evaluate the significance of each layer output. In this criteria, we rely on this \textit{scaling} coefficient as a proxy to the importance of each unit 
\begin{equation}
    \mathcal{C}_{norm}(\bW^{(l)}_{i})= \left| \mathbf{\gamma}^{(l)}_{i} \right|.
\end{equation}
Note that this criterion forces each layer to be followed by a normalization layer, from which it can be computed. 

\section{Experiments}

We briefly detail the tasks on which we evaluate our method for ultra-light deep MIR. As we address a wide variety of models and datasets, we only provide essential explanations for each. However, unless stated, we follow all implementation details presented in the original papers.

\subsection{Tasks}

\subsubsection{Audio (instrument and voice) classification} 
\textit{Audio classification} is one of the seminal and most studied task in MIR \cite{tzanetakis2002musical}. We separate the evaluation into two independent sub-tasks of \textit{singing voice} and \textit{instrument} classification. For both tasks, the model is adapted from the baseline proposed in \cite{wilkins2018vocalset}. The raw input waveform is processed with a stack of 4 dilated 1-dimensional convolutions with batch normalization, ReLU and dropout, followed by 4 fully-connected layers that map to a \textit{softmax}, which outputs a vector of class probabilities. The ground-truth label prediction is optimized with a \textit{cross-entropy} loss. Singing voice classification is performed on mono audio inputs of 3 seconds at 44,100Hz for 10 vocal techniques and a given train/test split ratio \cite{wilkins2018vocalset}. For instrument classification, we rely on the 13 orchestral instruments from URMP \cite{urmp} and the corresponding recordings from MedleyDB \cite{medleydb}. After silence removal, the combined datasets amount to a total of about 8h30 of isolated instrument recordings. Classification is done on audio inputs of 1.5 seconds at 22,050Hz extracted from the isolated tracks with a single label corresponding to the instrument played.

\subsubsection{Pitch estimation}

The goal of \textit{pitch estimation} is to extract the fundamental frequency of an input audio. For this task, the recently proposed CREPE model \cite{kim2018crepe} requires several large datasets, some of which are not publicly available. However, we only rely here on the open source \textit{NSynth} dataset, which contains single note samples from a range of acoustic and electronic instruments \cite{engel2017nsynth}. This leads to 1006 instruments, with different pitches at various velocities available as raw waveforms. All samples last 4 seconds with a sampling-rate of 16kHz. As this incurs an extremely large training time, we use a subsampled dataset, randomly picking 10060 samples (ten notes per instrument). Finally, we trim all samples to their first two seconds to remove silent note tails, ensuring that most inputs to the model are voiced.
CREPE is a 6-layer CNN operating directly on waveforms, followed by a single fully connected layer. The model is trained via binary cross-entropy to perform classification over a 360 bin logarithmic frequency scale spanning six octaves from pitch \( C1 \) to \( B7 \). The model operates on frames of 1024 samples, which we individually label with the note pitch. We use the \textit{medium} architecture from the CREPE repository~\cite{kim2018crepe}.

\subsubsection{Chord extraction}
\textit{Automatic chord extraction} is defined as labeling segments of an audio signal using an alphabet of musical chords. We perform our experiments based on the model and datasets detailed in \cite{carsault2018using}. We use the \emph{Beatles} dataset, which contains 180 songs annotated by hand. We rely on a CQT input with hop size 2048, mapped to a scale of 2 bins per semi-tone over 5 octaves starting from \( C1 \) and containing a total of 105 bins. As input we take 15 successive frames, corresponding to a temporal horizon of approximately 0.7 seconds. We augment the available data by performing all transpositions from -6 to +6 semi-tones. As baseline model, we rely on the CNN architecture described in \cite{carsault2018using}, and evaluate the global accuracy measure.

\begin{figure*}
 \includegraphics[width=1.02\textwidth]{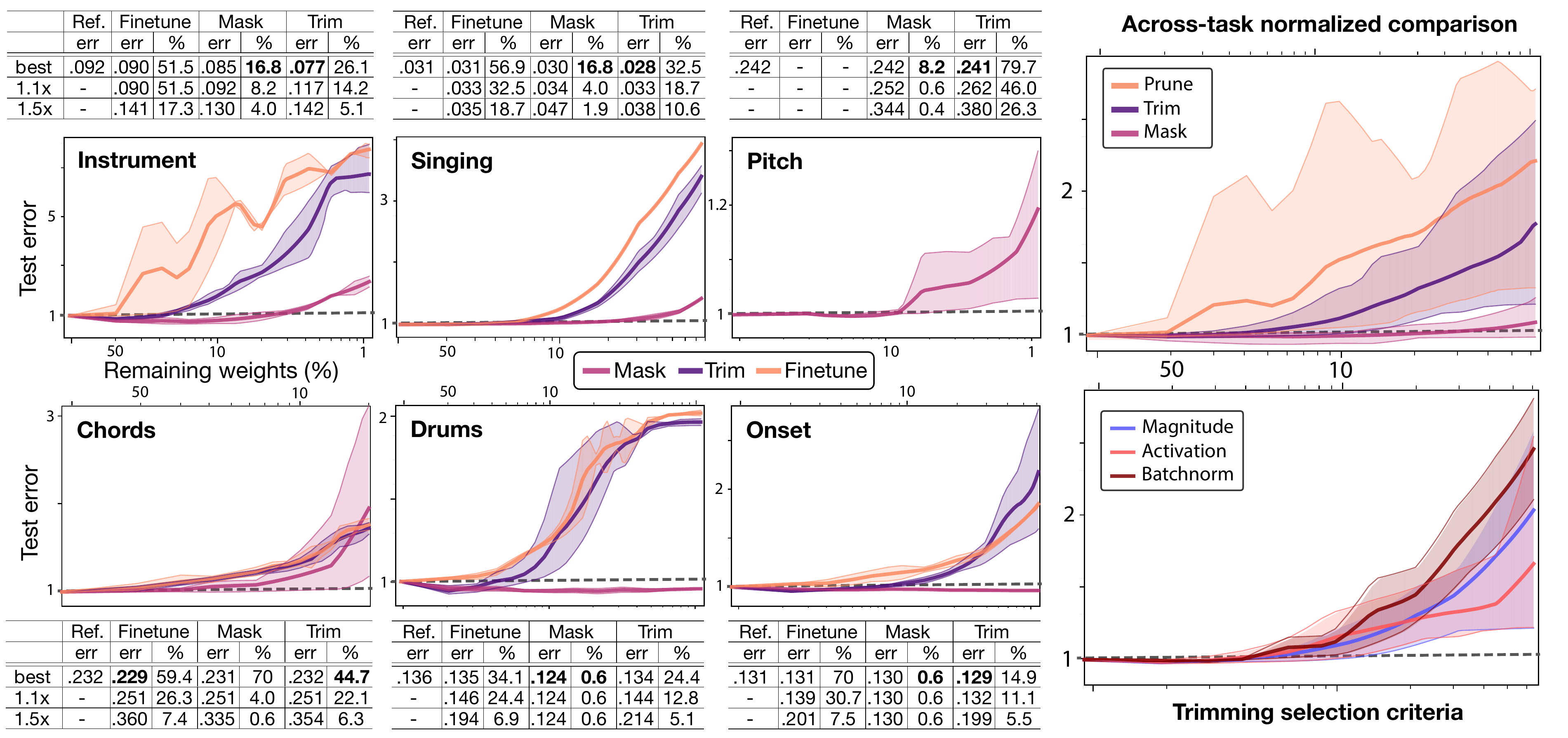}
 \caption{Comparing traditional \textit{fine-tuning}, the lottery ticket \textit{masking}, and our structured lottery \textit{trimming} on each MIR task separately for all criteria (left), and across tasks (right, up) and across criteria in trimming (right, down).}
 \label{fig:results}
\end{figure*}

\subsubsection{Drum transcription}

The \textit{drum transcription} task aims at labelling an input audio with onsets of different drum sounds. We rely on an architecture inspired from \cite{schluter2014improved}. The network takes mel-spectrogram inputs processed by 4 layers of 256 padded 2D convolutions of kernel size 5, with unit stride, followed by batch normalization and ReLU. We apply max-pooling at each layer along the frequency dimension. The resulting vector is processed by two linear layers with batch normalization and dropout. Finally, a specific output network of 3 linear layers for each drum sound produces onset probabilities, trained on binary vectors for each drum activation.

To train the network, we use the approach proposed by  \cite{cartwright2018increasing}, using a subset of 5000 MIDI drum tracks that we map to random drum sounds to generate waveform recordings. We further rely on the SMT-Drums dataset \cite{dittmar2014real}, which provides 104 supplementary polyphonic drum set recordings. For both datasets, we compute a mel-spectrogram of 64 bins, ranging from 20 to 11025 Hz, based on a FFT of window size 2048 and hop size 512.

\subsubsection{Onset estimation}

\textit{Onset estimation} \cite{schluter2014improved} aims to detect events in a given audio input. In order to evaluate this task, we rely on the same network presented in the previous section for drum transcription. However, for this task, the last part of the network maps to a single detection subnetwork. We rely on the same drums dataset, but merge all labels to detect event onsets, rather than specific elements of the drumkit.

\subsection{Training}

All models are trained following their respective procedure and hyperparameters. However, we use a common minibatch size of 64, the ADAM optimizer with a weight decay penalty of $2e^{-4}$, and an initial learning rate of $1e^{-3}$, which is halved every $10$ non-decreasing epochs. We train the models for a  number of epochs that is fixed for each task (following the original papers) and keep the model with the best validation score. For the lottery training, we perform \textit{masking} or \textit{trimming} of $30\%$ of the weights at each pruning iteration. We rewind the weights to their values at half of the training epochs. We repeat this process $15$ times, leading to models with up to $99.5\%$ of weights removed. This whole lottery training is repeated $5$ times, providing the variance and impact of the initialization on the results.

\section{Results}

\subsection{Global evaluation}

First, we provide a global evaluation across different tasks, by plotting the respective evolution of the test error as we iteratively remove weights either by classical \textit{fine-tuning}, using the original lottery with \textit{masking}, or our proposed \textit{trimming}. We report for each task the \textit{best} model (lowest test error), \textit{smallest} model (test error at most 1.5 times the original one), and \textit{optimal} model (error at most 1.1 times the original). Results are displayed in \figref{fig:results}.

As we can see, classical \textit{fine-tuning} is mostly unable to find more efficient lighter networks and only works at very low pruning rates. Oppositely, our \textit{trimming} approach is able to consistently find networks that are both much smaller and more accurate than reference models. In this regard, the best performances are obtained for \textit{onset detection}, where we find a network with only \textbf{14.9}\% of the original weights (removing \textbf{85.1}\% of the weights), while having an error rate of \textbf{0.129} (compared to 0.131 for the original). These results hold for almost all tasks: most networks where we trim up to \textbf{75}\% of the weights produce lower errors, and we can remove up to \textbf{85}\% of the weights with minor damage to the test error. Interestingly, the results of \textit{chord extraction} seems to produce the smallest enhancement. This could be explained by the fact that the model has the lowest original number of parameters. Hence, this underlines the crucial need to rely on \textit{largely overparametrized} models to find efficient subnetworks. Regarding the \textit{smallest} models, we are able to remove on average up to 95\%, while having a reasonable test error. When comparing our approach to the original lottery \textit{masking}, it seems that masking consistently produces better performances at higher pruning rates, confirming the original results \cite{frankle2018lottery} for MIR tasks. However, note that the weights in the masking approach are not removed (however, a fraction of these weights could be removed in a post-processing step). We hypothesize that this resilience to larger pruning ratios stems from the fact that \textit{masking} is able to work on local connectivity patterns, whereas our approach cannot.

\subsection{Across-task comparison} 

\subsubsection{Pruning approaches}

The lottery ticket hypothesis crucially depends on initialization values for training efficient subnetworks. To evaluate this property across different tasks, we perform the normalized comparison shown in \figref{fig:results} (right, up).

Here, we normalize the error of each task by dividing it by the error of the reference large model, so that its test error is 1. As we can see, using fine-tuning, the approach is unable to obtain subnetworks with higher accuracy, and the error quickly degrades as we remove more weights. Furthermore, it appears that the results are rather unstable, producing large variations in the final test error. Instead, by \textit{rewinding} the weights and \textit{trimming} we consistently obtain smaller subnetworks (up to 75\% of the weights removed) that outperform the original models. We are able to apply extensive trimming before the error starts to degrade, globally around 90\% across tasks. Hence, it appears that efficient subnetworks can be found solely through the correct combination of connection topology and weights. 



\subsubsection{Selection criteria}

The success of pruning methods depends on the criterion selecting \textit{which} weights should be kept or pruned. 
Hence, we perform a normalized comparison of different criteria for \textit{trimming}, and display results in \figref{fig:results} (right, down).

Although the global trend seems to be equivalent for most criteria at low pruning ratios, their differences amplify as we remove an increasing amount of weights. Overall, it seems that the \textit{activation} criterion provides the most stable results. Furthermore, it allows to maintain lower error rates, even at higher pruning ratios. However, at lower pruning ratios, it  seems that the \textit{magnitude} criterion produces slightly better and more stable results. Finally, the \textit{batchnorm} criterion seems to provide an interesting alternative at low pruning ratios. However, its performance degrades faster than other criteria at very high pruning rates.



\subsection{Resulting model properties}

We provide a detailed analysis of the gains provided by our \textit{trimming lottery} for each task. We compare the reference model to the \textit{optimal} one (smallest model within 1.1 times the original test error) found by trimming. We evaluate their \textit{test error}, \textit{number of parameters}, \textit{disk size}, \textit{FLOPS} (required to infer from a single input example) and \textit{memory used} for different MIR tasks, as detailed in \tabref{tab:tasks}. As discussed previously, we are able to obtain models maintaining the error rates, while having only a small portion of the capacity of the very large models. This can be witnessed in the final properties of the trimmed models. A very interesting observation is that this decrease in parameters amounts to an even larger decrease in the \textit{memory} and \textit{computation power} required. Indeed, while most trimmed models are 10 times smaller than original large models, they use 20 to 50 times less computation power and memory requirements. This can be explained by the fact that most operations are processed across the dimensions of the previous layer. Hence, even small gains in number of parameters can lead to dramatic gains in computation.


\begin{table}
\footnotesize
\begin{centering}
\begin{tabular}{c|c|c|c|c|c|c}
\hline 
\textbf{task} & \textbf{mod.} & \textbf{error} & \textbf{param} & \textbf{size} & \textbf{FLOPS} & \textbf{mem}\tabularnewline
\hline 
\hline 
\multirow{2}{*}{inst.} & ref & 0.092 & 797 K & 10 M & 572 M & 190 M\tabularnewline
 & trim & 0.117 & 93.4 K & 2.3 M & 38.3 M & 41.9 M\tabularnewline
\hline 
\multirow{2}{*}{sing.} & ref & 0.031 & 1.4 M & 19 M & 663 M & 194 M \tabularnewline
 & trim & 0.038 & 144 K & 2.7 M & 94.4 M & 53.2 M\tabularnewline
\hline 
\multirow{2}{*}{pitch} & ref & 0.242 & 5.9 M & 49 M & 2.8 G & 256 M\tabularnewline
 & trim & 0.262 & 224 K & 1.0 M & 2.8 M & 9.6 M\tabularnewline
\hline 
\multirow{2}{*}{chord} & ref & 0.232 & 416 K & 1.4 M & 27.2 M & 22.1 M \tabularnewline
 & trim & 0.251 & 91.9 K & 0.2 M & 1.72 M & 589 K\tabularnewline
\hline 
\multirow{2}{*}{drum} & ref & 0.136 & 8.1 M & 22 M & 3.54 G & 667 M\tabularnewline
 & trim & 0.144 & 1.0 M & 3.7 M & 87.5 M & 10.2 M\tabularnewline
\hline 
\multirow{2}{*}{onset} & ref & 0.131 & 4.7 M & 21 M & 2.66 G & 532 M\tabularnewline
 & trim & 0.132 & 522 K & 3.7 M & 87.1 M & 8.2 M\tabularnewline
\hline 
\end{tabular}
\par\end{centering}
\caption{Comparison between reference models and our trimmed models on \textit{test error}, \textit{number of parameters}, \textit{disk size}, \textit{inference FLOPS} and \textit{memory used} across tasks.}
\label{tab:tasks}
\end{table}

\section{Conclusion}

In this paper, we presented a method to obtain ultra-light deep models for MIR, by extending the lottery ticket hypothesis to effectively trim the networks. We have shown that these efficient trimmed subnetworks, removing up to 85\% of the weights in deep models, could be found across several MIR tasks. We have also shown that extremely small networks could be found by relying on \textit{masking}, but these do not provide actual enhancement in terms of computation or memory requirements. Oppositely, we have shown that given the non-linear relationship between the number of parameters and computation required, we could find extremely light networks through trimming. These results encourage the crucial implementation of MIR models in embedded audio platforms, which would allow broader end-user applications. The major downside of this approach is its training time, which we partly address by decreasing the cost of each pruning iteration. However, the intriguing prospect of \textit{ticket transfer} \cite{morcos2019one} could provide such initializations right \emph{at the onset} of training. 

\section{Acknowledgements}

This work is supported by the ANR:17-CE38-0015-01 MAKIMOno project, the SSHRC:895-2018-1023 ACTOR Partnership and Emergence(s) ACIDITEAM project from Ville de Paris and ACIMO projet of Sorbonne Universit\'e.

\bibliography{main}

\begin{thebibliography}{10}
\providecommand{\url}[1]{#1}
\csname url@samestyle\endcsname
\providecommand{\newblock}{\relax}
\providecommand{\bibinfo}[2]{#2}
\providecommand{\BIBentrySTDinterwordspacing}{\spaceskip=0pt\relax}
\providecommand{\BIBentryALTinterwordstretchfactor}{4}
\providecommand{\BIBentryALTinterwordspacing}{\spaceskip=\fontdimen2\font plus
\BIBentryALTinterwordstretchfactor\fontdimen3\font minus
  \fontdimen4\font\relax}
\providecommand{\BIBforeignlanguage}[2]{{%
\expandafter\ifx\csname l@#1\endcsname\relax
\typeout{** WARNING: IEEEtran.bst: No hyphenation pattern has been}%
\typeout{** loaded for the language `#1'. Using the pattern for}%
\typeout{** the default language instead.}%
\else
\language=\csname l@#1\endcsname
\fi
#2}}
\providecommand{\BIBdecl}{\relax}
\BIBdecl

\bibitem{casey2008content}
M.~A. Casey, R.~Veltkamp, M.~Goto, M.~Leman, C.~Rhodes, and M.~Slaney,
  ``Content-based music information retrieval: Current directions and future
  challenges,'' \emph{Proceedings of the IEEE}, vol.~96, no.~4, pp. 668--696,
  2008.

\bibitem{humphrey2012moving}
E.~J. Humphrey, J.~P. Bello, and Y.~LeCun, ``Moving beyond feature design: Deep
  architectures and automatic feature learning in music informatics.'' in
  \emph{12th International Society for Music Information Retrieval Conference
  (ISMIR)}, 2012, pp. 403--408.

\bibitem{you2017imagenet}
Y.~You, Z.~Zhang, C.-J. Hsieh, J.~Demmel, and K.~Keutzer, ``{ImageNet} training
  in minutes,'' in \emph{Proceedings of the 47th International Conference on
  Parallel Processing}, 2018, pp. 1--10.

\bibitem{he2016deep}
K.~He, X.~Zhang, S.~Ren, and J.~Sun, ``Deep residual learning for image
  recognition,'' in \emph{Proceedings of the IEEE conference on Computer Vision
  and Pattern Recognition (CVPR)}, 2016, pp. 770--778.

\bibitem{lecun1990optimal}
Y.~LeCun, J.~S. Denker, and S.~A. Solla, ``Optimal brain damage,'' in
  \emph{Advances in Neural Information Processing Systems (NIPS)}, 1990, pp.
  598--605.

\bibitem{han2015deep}
S.~Han, H.~Mao, and W.~J. Dally, ``Deep compression: Compressing deep neural
  networks with pruning, trained quantization and huffman coding,'' \emph{arXiv
  preprint arXiv:1510.00149}, 2015.

\bibitem{hinton2015distilling}
G.~Hinton, O.~Vinyals, and J.~Dean, ``Distilling the knowledge in a neural
  network,'' in \emph{NIPS Deep Learning and Representation Learning Workshop},
  2015.

\bibitem{liu2018rethinking}
Z.~Liu, M.~Sun, T.~Zhou, G.~Huang, and T.~Darrell, ``Rethinking the value of
  network pruning,'' in \emph{International Conference on Learning
  Representations}, 2019.

\bibitem{gale2019state}
T.~Gale, E.~Elsen, and S.~Hooker, ``The state of sparsity in deep neural
  networks,'' \emph{arXiv preprint arXiv:1902.09574}, 2019.

\bibitem{frankle2018lottery}
J.~Frankle and M.~Carbin, ``The lottery ticket hypothesis: Finding sparse,
  trainable neural networks,'' in \emph{International Conference on Learning
  Representations (ICLR)}, 2019.

\bibitem{frankle2019linear}
J.~Frankle, G.~K. Dziugaite, D.~M. Roy, and M.~Carbin, ``Linear mode
  connectivity and the lottery ticket hypothesis,'' \emph{arXiv preprint
  arXiv:1912.05671}, 2019.

\bibitem{frankle2019lottery}
------, ``Stabilizing the lottery ticket hypothesis,'' \emph{arXiv preprint
  arXiv:1903.01611}, 2019.

\bibitem{frankle2020the}
J.~Frankle, D.~J. Schwab, and A.~S. Morcos, ``The early phase of neural network
  training,'' in \emph{International Conference on Learning Representations},
  2020.

\bibitem{morcos2019one}
A.~Morcos, H.~Yu, M.~Paganini, and Y.~Tian, ``One ticket to win them all:
  generalizing lottery ticket initializations across datasets and optimizers,''
  in \emph{Advances in Neural Information Processing Systems (NIPS)}, 2019, pp.
  4933--4943.

\bibitem{you2019drawing}
H.~You, C.~Li, P.~Xu, Y.~Fu, Y.~Wang, X.~Chen, R.~G. Baraniuk, Z.~Wang, and
  Y.~Lin, ``Drawing early-bird tickets: Toward more efficient training of deep
  networks,'' in \emph{International Conference on Learning Representations
  (ICLR)}, 2019.

\bibitem{esling2018bridging}
P.~Esling, A.~Chemla-Romeu-Santos, and A.~Bitton, ``Bridging audio analysis,
  perception and synthesis with perceptually-regularized variational timbre
  spaces.'' in \emph{18th International Society for Music Information Retrieval
  Conference (ISMIR)}, 2018.

\bibitem{wilkins2018vocalset}
J.~Wilkins, P.~Seetharaman, A.~Wahl, and B.~A. Pardo, ``{VocalSet}: A singing
  voice dataset,'' in \emph{{Proceedings of the 19th International Society for
  Music Information Retrieval (ISMIR) Conference}}, 2018, pp. 468--474.

\bibitem{kim2018crepe}
J.~W. Kim, J.~Salamon, P.~Li, and J.~P. Bello, ``Crepe: A convolutional
  representation for pitch estimation,'' in \emph{2018 IEEE International
  Conference on Acoustics, Speech and Signal Processing (ICASSP)}.\hskip 1em
  plus 0.5em minus 0.4em\relax IEEE, 2018, pp. 161--165.

\bibitem{carsault2018using}
T.~Carsault, J.~Nika, and P.~Esling, ``Using musical relationships between
  chord labels in automatic chord extraction tasks,'' in \emph{International
  Society for Music Information Retrieval (ISMIR) Conference}, 2018.

\bibitem{choi2019deep}
K.~Choi and K.~Cho, ``Deep unsupervised drum transcription,'' in \emph{20th
  International Society for Music Information Retrieval (IISMIR) Conference},
  2019.

\bibitem{schluter2014improved}
J.~Schl{\"u}ter and S.~B{\"o}ck, ``Improved musical onset detection with
  convolutional neural networks,'' in \emph{IEEE International Conference on
  Acoustics, Speech and Signal Processing (ICASSP)}, 2014, pp. 6979--6983.

\bibitem{han2015learning}
S.~Han, J.~Pool, J.~Tran, and W.~Dally, ``Learning both weights and connections
  for efficient neural network,'' in \emph{Advances in Neural Information
  Processing Systems (NIPS)}, 2015, pp. 1135--1143.

\bibitem{li2016pruning}
H.~Li, A.~Kadav, I.~Durdanovic, H.~Samet, and H.~P. Graf, ``Pruning filters for
  efficient {ConvNets},'' in \emph{International Conference on Learning
  Representations (ICLR)}, 2017.

\bibitem{shwartz2017opening}
R.~Shwartz-Ziv and N.~Tishby, ``Opening the black box of deep neural networks
  via information,'' \emph{arXiv preprint arXiv:1703.00810}, 2017.

\bibitem{mcfee2012learning}
B.~McFee, L.~Barrington, and G.~Lanckriet, ``Learning content similarity for
  music recommendation,'' \emph{IEEE Transactions on Audio, Speech, and
  Language Processing}, vol.~20, no.~8, pp. 2207--2218, 2012.

\bibitem{humphrey2013feature}
E.~J. Humphrey, J.~P. Bello, and Y.~LeCun, ``Feature learning and deep
  architectures: New directions for music informatics,'' \emph{Journal of
  Intelligent Information Systems}, vol.~41, no.~3, pp. 461--481, 2013.

\bibitem{esling2020flow}
P.~Esling, N.~Masuda, A.~Bardet, R.~Despres, and A.~Chemla-Romeu-Santos, ``Flow
  synthesizer: Universal audio synthesizer control with normalizing flows,''
  \emph{Applied Sciences}, vol.~10, no.~1, p. 302, 2020.

\bibitem{choi2017tutorial}
K.~Choi, G.~Fazekas, K.~Cho, and M.~Sandler, ``A tutorial on deep learning for
  music information retrieval,'' \emph{arXiv preprint arXiv:1709.04396}, 2017.

\bibitem{de2002yin}
A.~De~Cheveign{\'e} and H.~Kawahara, ``Yin, a fundamental frequency estimator
  for speech and music,'' \emph{The Journal of the Acoustical Society of
  America}, vol. 111, no.~4, pp. 1917--1930, 2002.

\bibitem{tzanetakis2002musical}
G.~Tzanetakis and P.~Cook, ``Musical genre classification of audio signals,''
  \emph{IEEE Transactions on Audio, Speech, and Language Processing}, vol.~10,
  no.~5, pp. 293--302, 2002.

\bibitem{urmp}
B.~Li, X.~Liu, K.~Dinesh, Z.~Duan, and G.~Sharma, ``Creating a multitrack
  classical music performance dataset for multimodal music analysis:
  Challenges, insights, and applications,'' \emph{IEEE Transactions on
  Multimedia}, vol.~21, no.~2, pp. 522--535, 2019.

\bibitem{medleydb}
R.~Bittner, J.~Salamon, M.~Tierney, M.~Mauch, C.~Cannam, and J.~Bello,
  ``Medleydb: A multitrack dataset for annotation-intensive mir research,'' in
  \emph{Proceedings of the 15th International Society for Music Information
  Retrieval (ISMIR) Conference}, 2014.

\bibitem{engel2017nsynth}
J.~Engel, C.~Resnick, A.~Roberts, S.~Dieleman, D.~Eck, K.~Simonyan, and
  M.~Norouzi, ``Neural audio synthesis of musical notes with {WaveNet}
  autoencoders,'' \emph{International Conference on Machine Learning}, vol.~70,
  pp. 1068--1077, 2017.

\bibitem{cartwright2018increasing}
M.~Cartwright and J.~P. Bello, ``Increasing drum transcription vocabulary using
  data synthesis,'' in \emph{Proc. International Conference on Digital Audio
  Effects (DAFx)}, 2018, pp. 72--79.

\bibitem{dittmar2014real}
C.~Dittmar and D.~G{\"a}rtner, ``Real-time transcription and separation of drum
  recordings based on {NMF} decomposition.'' in \emph{Proc. International
  Conference on Digital Audio Effects (DAFx)}, 2014, pp. 187--194.

\end{thebibliography}

\end{document}